\pdfoutput=1

\documentclass[11pt]{article}

\usepackage{EACL2023}

\usepackage{times}
\usepackage{latexsym}

\usepackage[T1]{fontenc}

\usepackage[utf8]{inputenc}

\usepackage{microtype}

\usepackage{inconsolata}

\usepackage{multirow}
\newcommand{\bhline}{\noalign{\hrule height 1.2pt}}
\usepackage{arydshln}
\usepackage{graphicx}
\usepackage{amssymb}
\usepackage{amsmath}
\usepackage{CJKutf8}
\usepackage{bm}

\title{Semantic Frame Induction with Deep Metric Learning}

\author{
  Kosuke Yamada$^{1}$ \ \ \ \ \ \ \ \ \ 
  Ryohei Sasano$^{1,2}$ \ \ \ \ \ \ \ \ \ 
  Koichi Takeda$^{1}$  \\
  $^{1}$Graduate School of Informatics, Nagoya University, Japan \\
  $^{2}$RIKEN Center for Advanced Intelligence Project, Japan \\
  {\tt yamada.kosuke.v1@s.mail.nagoya-u.ac.jp}, \\
  {\tt \{sasano,takedasu\}@i.nagoya-u.ac.jp}
}

\begin{document}
\maketitle
\begin{abstract}
Recent studies have demonstrated the usefulness of contextualized word embeddings in unsupervised semantic frame induction. 
However, they have also revealed that generic contextualized embeddings are not always consistent with human intuitions about semantic frames, which causes unsatisfactory performance for frame induction based on contextualized embeddings.
In this paper, we address \textit{supervised} semantic frame induction, which assumes the existence of frame-annotated data for a subset of predicates in a corpus and aims to build a frame induction model that leverages the annotated data.
We propose a model that uses deep metric learning to fine-tune a contextualized embedding model, and we apply the fine-tuned contextualized embeddings to perform semantic frame induction.
Our experiments on FrameNet show that fine-tuning with deep metric learning considerably improves the clustering evaluation scores, namely, the \textsc{B-cubed F-score} and \textsc{Purity F-score}, by about 8 points or more.
We also demonstrate that our approach is effective even when the number of training instances is small.
\end{abstract}

\section{Introduction}
Semantic frames are knowledge resources that reflect human intuitions about various concepts such as situations and events.
One of the most representative semantic frame resources is FrameNet \cite{baker1998,ruppenhofer2016}, which consists of semantic frames, lexical units (LUs) that evoke these frames, and collections of frame-annotated sentences.
Semantic frame induction is the task of grouping predicates, typically verbs, according to the semantic frames they evoke.\footnote{Strictly speaking, this task can be divided into two sub-tasks: verb clustering, which groups verbs according to the frames that they evoke, and argument clustering, which groups arguments of verbs according to their roles.
In this study, we focus the former, the verb clustering task.}
For example, given the verbs in the example sentences listed in Table \ref{tab:examples}, semantic frame induction aims to group them into four clusters according to the frames that they evoke.

\begin{table}[!t]
\small
\centering
\begin{tabular}{@{}l@{\ }l@{}} \bhline
\rule{0ex}{2.0ex}
Frame & Example sentence \\ \hline
\rule{0ex}{2.0ex}
\multirow{2}{*}{\color[HTML]{1f77b4} \textsc{Filling}} 
& (1) She {\color[HTML]{1f77b4} covered} her mouth with her hand.  \\
\rule{0ex}{1.0ex}
& (2) I {\color[HTML]{1f77b4} filled} a notebook with my name. \\\hdashline
\rule{0ex}{2.0ex}
\multirow{2}{*}{\color[HTML]{ff7f0e} \textsc{Placing}} 
& (3) You can {\color[HTML]{ff7f0e} embed} graphs in your worksheet.  \\
\rule{0ex}{1.0ex}
& (4) He {\color[HTML]{ff7f0e} parked} the car at the hotel. \\\hdashline
\rule{0ex}{2.0ex}
\multirow{2}{*}{\color[HTML]{2ca02c} \textsc{Removing}}
& (5) Volunteers {\color[HTML]{2ca02c} removed} grass from the marsh. \\
\rule{0ex}{1.0ex}
& (6) They'd {\color[HTML]{2ca02c} drained} the drop from the teapot. \\\hdashline
\rule{0ex}{2.0ex}
\multirow{2}{*}{\color[HTML]{d62728} \textsc{Topic}} 
& (7) Each database will {\color[HTML]{d62728} cover} a specific topic.  \\
\rule{0ex}{1.0ex}
& (8) Chapter 8 {\color[HTML]{d62728} treats} the educational advantages. \\ \bhline
\end{tabular}
\caption{Example sentences and the frames that their verbs evoke in FrameNet.}
\label{tab:examples}
\end{table}

\begin{figure}[!t]
\centering
\includegraphics[width=\linewidth]{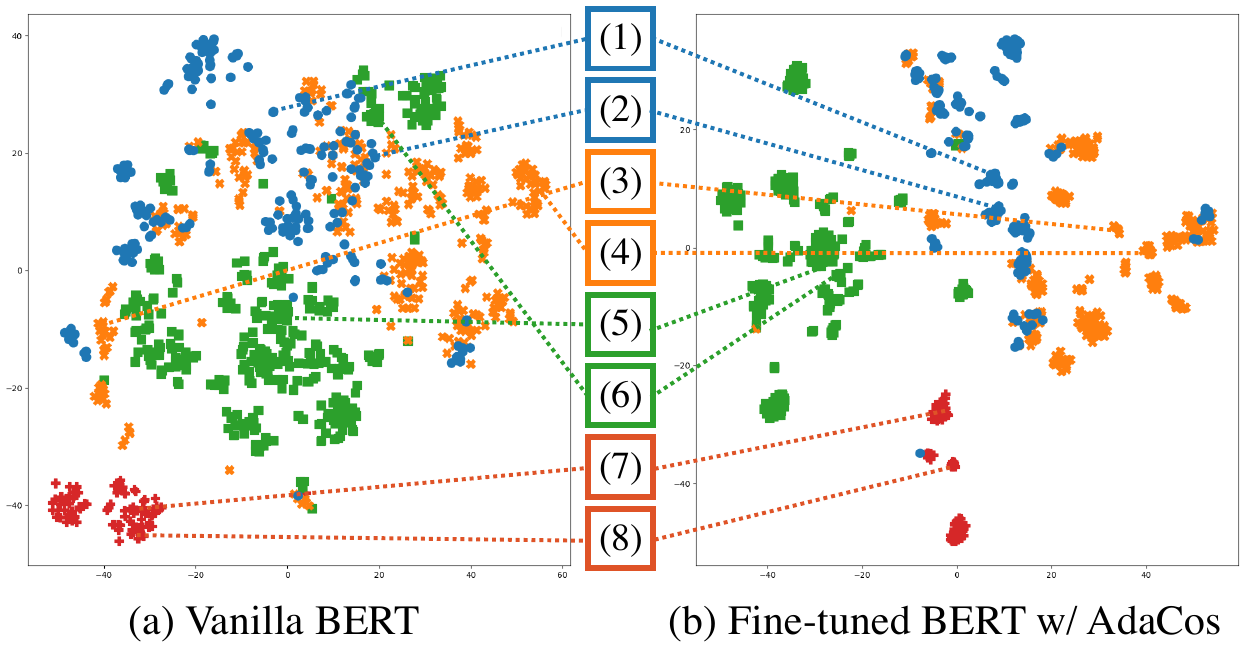}
\caption{2D t-SNE projections of the contextualized embeddings of verbs annotated with either the {\color[HTML]{1f77b4} \textsc{Filling}} ({\color[HTML]{1f77b4} \large $\bullet$}), {\color[HTML]{ff7f0e} \textsc{Placing}} ({\color[HTML]{ff7f0e} $\bm \times$}), {\color[HTML]{2ca02c} \textsc{Removing}} ({\color[HTML]{2ca02c} \scriptsize $\blacksquare$}), or {\color[HTML]{d62728} \textsc{Topic}} ({\color[HTML]{d62728} \textbf{+}}) frames in FrameNet, by using (a) vanilla BERT or (b) fine-tuned BERT with AdaCos.
The numbers in parentheses correspond to the examples listed in Table \ref{tab:examples}.} 
\label{fig:examples}
\end{figure}

Recent studies \cite{arefyev2019,anwar2019,ribeiro2019} have demonstrated the usefulness of contextualized word embeddings such as ELMo \cite{peters2018} and BERT \cite{devlin2019} in unsupervised semantic frame induction. 
Figure \ref{fig:examples}(a) shows a 2D t-SNE \cite{maaten2008} projection of the vanilla BERT\footnote{We refer to BERT without fine-tuning as ``vanilla BERT.''} embeddings of verbs extracted from frame-annotated sentences in FrameNet.
We can confirm that the instances of the verb ``cover'' in Examples (1) and (7) are far apart in the space, which reflects the distance between their meanings.
In contrast, ``cover'' in (7) and ``treat'' in (8), which are annotated with the same \textsc{Topic} frame, are close together.
However, instances are not always properly placed in the semantic space according to the frames they evoke.
For example, ``remove'' in (5) and ``drain'' in (6) are annotated with the same \textsc{Removing} frame but are not close together.
This suggests that contextualized word embeddings are not always consistent with human intuition about semantic frames.

Hence, in this study, we tackle \textit{supervised} semantic frame induction, which assumes the existence of annotated data for certain predicates, to induce semantic frames that adequately reflect human intuition about the frames.
We propose methods that use deep metric learning to fine-tune the contextual word embedding model so that instances of verbs that evoke the same frame are placed close together and other instances are placed farther apart in the semantic space.
Figure \ref{fig:examples}(b) shows the 2D t-SNE projection of BERT embeddings after fine-tuning with AdaCos \cite{zhang2019adacos}, which is a representative deep metric learning method.
We can confirm that predicates that evoke the same frame are close together, such as those in (3) and (4) and those in (5) and (6).
This suggests that deep metric learning enables fine-tuning of BERT to obtain embedding spaces that better reflect human intuition about semantic frames.

\section{Related Work}
For automatic construction of semantic frame resources, studies on grouping predicates according to the semantic frames they evoke can be divided into two groups: those that work on \textbf{semantic frame identification}, in which predicates are classified into predefined frames; and those that work on \textbf{semantic frame induction}, in which predicates are grouped according to the frames that they evoke, which are typically not given in advance.

Semantic frame identification is often treated as a subtask of frame semantic parsing \cite{das-etal-2014-frame,swayamdipta2017frame}, and the methods using contextualized embedding have become mainstream.
For example, \newcite{jiang-riloff-2021-exploiting} used a BERT-based model to generate representations for frames and LUs by using their formal definitions.
\newcite{su-etal-2021-knowledge} used a BERT-based model with a context encoder, to encode the context surrounding the frame-evoking word, and a frame encoder, to encode the frames' definitions and semantic roles.
\newcite{yong2020semi} treated semantic frame identification as a clustering task.\footnote{
Although they called their task semantic frame induction, 
we refer to the task as frame identification mainly due to the fact that their experiments excluded predicates that evoke unknown frames.}
They first excluded predicates that evoke frames that are not included in FrameNet by applying an anomaly detection model; then, they grouped the remaining predicates according to their meanings by using contextualized embeddings of predicates and sentence embeddings of the frame definitions.

Semantic frame induction is the task of grouping predicates in texts according to the frames they evoke.
Instead of frames being given in advance, each grouping of given predicates is considered a frame.
As with semantic frame identification, methods using contextualized embedding have become mainstream.
\newcite{arefyev2019} first performed agglomerative clustering by using the BERT embedding of a frame-evoking verb and then split each cluster into two on the basis of the verb's substitutes.
\newcite{anwar2019} used the embedding of the frame-evoking verb and the average word embedding of all the words in a sentence, as obtained by skip-gram \cite{mikolov2013} or ELMo and then performed agglomerative clustering.
\newcite{ribeiro2019} applied graph clustering by using the ELMo embedding of the frame-evoking verb.
\newcite{yamada-etal-2021-semantic} leveraged the embedding of the masked frame-evoking verb and performed two-step clustering, which comprised intra-verb and cross-verb clustering.
\newcite{yamada-etal-2021-verb} investigated how well contextualized word representations can recognize the
difference of frames that the same verb evokes, and explored which types of representation are suitable for semantic frame induction.
All of these studies focused on \textit{unsupervised} semantic frame induction, with no training data.
In contrast, in this study, we assume the existence of frame-annotated data for a subset of predicates appearing in a corpus, and work on \textit{supervised} semantic frame induction.

\section{Supervised Semantic Frame Induction}
\subsection{Task Description}
The task of supervised semantic frame induction assumes the existence of frame-annotated data for a subset of a corpus's predicates, and it aims to build a frame induction model that leverages the annotated data.
Clustering-based methods are generally used for semantic frame induction, and this is also true for supervised semantic frame induction, where the annotated data is used to learn the distance metric for clustering.
In this study, the predicates that are used for training the metric and for testing do not overlap.
Note that, because different predicates may evoke the same frame, instances in the test data include predicates that evoke frames that are present in the training data.

\subsection{Baseline Methods}
For the simplest baseline, we use a one-step clustering-based method with contextualized embedding.
The clustering method is group-average clustering based on the Euclidean distance.
We also leverage the masked word embeddings and two-step clustering proposed by \newcite{yamada-etal-2021-semantic}.
Regarding the former, we use a weighted average embedding ($v_{w+m}$) of the standard contextualized embedding of the frame-evoking word ($v_{word}$) and the masked word embedding ($v_{mask}$), which is a contextualized embedding of the frame-evoking word replaced by a special token ``[MASK].''
The embedding $v_{w+m}$ is defined using a weight parameter $\alpha$ as follows:
\begin{equation}
v_{w+m} = (1-\alpha) \cdot v_{word} + \alpha \cdot v_{mask}.
\label{eq:vwm}
\end{equation}

Two-step clustering performs clustering for each frame-evoking word with the same lemma,\footnote{Following \newcite{yamada-etal-2021-semantic}, we refer to each cluster generated by clustering in the first step as a pseudo-LU (pLU).} and it performs clustering over different frame-evoking words in the second step.
We use X-means \cite{pelleg2000} for the first step and group-average clustering based on the Euclidean distance for the second step. 
All other settings here are the same as in \newcite{yamada-etal-2021-semantic}.

\subsection{Fine-tuning by Deep Metric Learning}
For supervised semantic frame induction, we fine-tune contextualized word embedding models by applying deep metric learning \cite{kaya2019deep,musgrave2020metric} so that the instances of predicates that evoke the same frame are closer together and those of predicates that evoke different frames are further apart.
We apply two representative deep metric learning approaches: a distance-based approach and a classification-based approach.

\paragraph{Distance-based Approach}
This is a classical deep metric learning approach, and the models typically use multiple encoders to train the distance between a pair of instances.
In this approach, we use two losses, a contrastive loss and a triplet loss, to build frame induction models.

The contrastive loss \cite{hadsell2006dimensionality} is used to train the distance between a pair of instances by using a network of two encoders with shared parameters.
Specifically, the model is trained to keep instances of the same class close together and instances of different classes separated by a certain margin.
The loss is defined as follows:
\begin{equation}
\!\! L_{\mathrm{cont}} \! = \!
\begin{cases}
D \left(\boldsymbol{x}_i, \boldsymbol{x}_j \right) & i \! = \! j \\
\max \left(m \! - \! D\left(\boldsymbol{x}_i, \boldsymbol{x}_j \right), 0 \right) & i \! \neq \! j
\end{cases}, \!
\label{eq:contrastive}
\end{equation}
where $\boldsymbol{x}_i$ denotes an embedding of an instance belonging to the $i$-th class, $m$ denotes a margin, and $D$ denotes a distance function, which is generally the squared Euclidean distance.

The triplet loss \cite{weinberger2009distance} is used for training such that, for a triplet of instances, the distance between the anchor and negative instances, which are from different classes, is more than a certain margin greater than the distance between the anchor and positive instances, which are from the same class.
The loss is defined as follows:
\begin{equation}
\!\! L_{\mathrm{tri}} \! = \!
\max \left(D\left(\boldsymbol{x}_a, \boldsymbol{x}_p \right)
\! - \! D\left(\boldsymbol{x}_a, \boldsymbol{x}_n \right)
\! + \! m,
0 \right), \!
\label{eq:triplet}
\end{equation}
where $\boldsymbol{x}_a$, $\boldsymbol{x}_p$, and $\boldsymbol{x}_n$ denote embeddings of the anchor, positive, and negative instances, respectively, and $m$ and $D$ are the same as in Equation (\ref{eq:contrastive}).

We create pairs for each instance in the training set by randomly selecting instances of predicates that evoke the same frame as positives and instances of predicates that evoke different frames as negatives.
The margin to keep the negatives away is determined by the development set.

\paragraph{Classification-based Approach}
This is an approach that has recently become the standard for face recognition.
It basically uses a network that has an encoder to obtain instance embeddings and a linear layer for multiclass classification.
This is superior to the distance-based approach in that it does not require a sampling algorithm and saves memory because it uses only a single encoder.
The loss function is based on the softmax loss:
\begin{equation}
L_{\mathrm{soft}} =
- \log \frac{e^{\boldsymbol{w}_i^{\top} \boldsymbol{x}_i + b_i}}{\sum_{j=1}^{n} e^{\boldsymbol{w}_j^{\top} \boldsymbol{x}_i + b_j}},
\label{eq:softmax}
\end{equation}
where $\boldsymbol{x}_i$, $\boldsymbol{w}_i$, and $b_i$ denote an embedding of the instance, the linear layer's weight, and a bias term, respectively, for the $i$-th class, and $n$ denotes the number of classes.

Many losses used in face recognition have been adjusted by introducing different margins for the softmax loss \cite{liu2017sphereface,wang2018cosface,deng2019arcface}.
These losses typically remove the bias term $b_i$ of the softmax loss and transform the logit as $\boldsymbol{w}_i^{\top} \boldsymbol{x}_i = ||\boldsymbol{w}_i|| \cdot ||\boldsymbol{x}_i|| \cdot \cos \theta_i$, where $\theta_i$ is the angle between $\boldsymbol{w}_i$ and $\boldsymbol{x}_i$.
ArcFace \cite{deng2019arcface} has become a popular choice because of its superior geometric interpretation. 
It applies $l_2$ regularization to $\boldsymbol{w}_i$ and $\boldsymbol{x}_i$ and introduces an angular margin $m$ and a feature scale $s$ as hyperparameters to simultaneously enhance the intra-class compactness and inter-class discrepancy. 
The ArcFace loss is defined as follows:
\begin{equation}
\!\!\! L_{\mathrm{arc}} \! = \!
- \log \frac{e^{s \cdot \cos \left(\theta_i+m \right)}}
{e^{s \cdot \cos \left(\theta_i+m \right)}
\! + \! \sum_{j=1, j \neq i}^{n} e^{s \cdot \cos \theta_j}}.
\end{equation}

\newcite{zhang2019adacos} pointed out that the performance of these losses depends on the hyperparameters and they observed the behaviors of the angular margin and the feature scale.
As a result, they proposed the hyperparameter-free AdaCos loss, which removes the margin and applies the scale dynamically.
The AdaCos loss is defined as follows:
\begin{equation}
L_{\mathrm{ada}} =
- \log \frac{e^{\tilde{s} \cdot \cos \theta_i}}
{\sum_{j=1}^{n} e^{\tilde{s} \cdot \cos \theta_j}},
\end{equation}
where $\tilde{s}$ denotes the automatically tuned scale.

While the softmax and AdaCos losses do not require a hyperparameter search, ArcFace requires hyperparameters for the margin and feature scale.
Here, we explore only the margin because \newcite{zhang2019adacos} showed that the behavior of the margin and the scale are similar and the distance-based approach explores the margin.

\section{Experiment}
\label{sec:experiment}
To evaluate the usefulness of fine-tuning with deep metric learning, we experimented with supervised semantic frame induction by comparing previous non-fine-tuned models to various fine-tuned models ranging from typical to evolved ones.
By varying the number of training instances, we also verified that our models were effective even for training a small number of instances.

\subsection{Settings}
\paragraph{Dataset}
The dataset in our experiments was created by extracting example sentences in which the frame-evoking word was a verb from the FrameNet 1.7 dataset.\footnote{\url{https://framenet.icsi.berkeley.edu/}}
These example sentences were split into three sets such that sentences with the same verb were in the same set.
The proportions of polysemous verbs were equal.
We performed three-fold cross-validation with the three sets as the training, development, and test sets.
Table \ref{tab:dataset} lists the dataset statistics.
Note that the verbs, LUs, and instances did not overlap among the sets, but the frames did overlap.
The training set was used to fine-tune the contextualized word embeddings. 
The development set was used to determine the criterion for the number of clusters and the weight $\alpha$ of the embedding $v_{\textsc{w+m}}$, as well as the margin for the contrastive, triplet, and ArcFace losses.
The range of $\alpha$ was from 0 to 1 in increments of 0.1, and the candidates of the margin were 0.1, 0.2, 0.5, and 1.0 for the contrastive and triplet losses and 0.01, 0.02, 0.05, and 0.1 for the ArcFace loss.

\begin{table}[!t]
\centering
\begin{tabular}{l@{\ }rrrr} \bhline
& \multicolumn{1}{c}{\#Verbs}  & \multicolumn{1}{c}{\#LUs} & \multicolumn{1}{c}{\#Frames} & \multicolumn{1}{c}{\#Instances} \\ \hline
Set 1   & 831   & 1,277 & 429 & 28,314 \\ 
Set 2   & 831   & 1,261 & 415 & 26,179 \\ 
Set 3   & 830   & 1,280 & 459 & 28,117 \\ \hline
All     & 2,492 & 3,818 & 642 & 82,610 \\ \bhline
\end{tabular}
\caption{Statistics of the FrameNet-based dataset used in three-fold cross-validation.}
\label{tab:dataset}
\end{table}

\begin{table*}[t!]
\centering
\begin{tabular}{cc crccc} \bhline
Clustering & Model & $\alpha$ 
 & \multicolumn{1}{c}{\#pLU} & \#C & \textsc{Pu} / \textsc{iPu} / \textsc{PiF} 
 & \textsc{BcP} / \textsc{BcR} / \textsc{BcF} \\ \hline

& Vanilla       & 0.00 & -- & 429 & 53.0 / 57.0 / 54.9 & 40.8 / 44.6 / 42.6 \\ \cdashline{2-7}
& Contrastive   & 0.13 & -- & 443 & 56.9 / 70.0 / 62.8 & 45.1 / 58.6 / 51.0 \\
One-step clustering
& Triplet       & 0.23 & -- & 425 & 70.0 / 77.0 / 73.3 & \textbf{60.3} / 68.1 / \textbf{63.9} \\
{\small (group-average clustering)} 
& Softmax       & 0.23 & -- & 440 & 65.1 / 78.0 / 71.0 & 53.3 / 68.6 / 59.9 \\
& ArcFace       & 0.37 & -- & 436 & \textbf{70.3} / 76.2 / 73.1 & 59.7 / 67.4 / 63.3 \\
& AdaCos        & 0.30 & -- & 446 & 69.0 / \textbf{78.7} / \textbf{73.5} & 57.5 / \textbf{69.5} / 62.9 \\ \hline \hline

& Vanilla       & 0.67 & 877    & 444 & 60.6 / 74.9 / 66.9 & 49.7 / 65.8 / 56.5 \\ \cdashline{2-7}
\multirow{2}{*}{Two-step clustering}
& Contrastive   & 0.23 & 1904   & 689 & 69.2 / 62.5 / 65.7 & 59.5 / 50.9 / 54.8 \\
\multirow{2}{*}{{\small (X-means \&}}
& Triplet       & 0.50 & 1014   & 454 & 73.4 / \textbf{76.7} / 74.8 & 64.6 / \textbf{68.0} / 66.0 \\
\multirow{2}{*}{\small group-average clustering)}
& Softmax       & 0.43 & 1428   & 919 & \textbf{84.7} / 62.5 / 71.9 & \textbf{78.4} / 50.4 / 61.4 \\
& ArcFace       & 0.47 & 955    & 452 & 70.5 / 76.5 / 73.3 & 60.8 / 67.7 / 63.8 \\
& AdaCos        & 0.50 & 1128   & 656 & 80.8 / 71.3 / \textbf{75.6} & 73.2 / 60.9 / \textbf{66.2} \\ \bhline
\end{tabular}
\caption{Experimental results on semantic frame induction with vanilla and fine-tuned models over three-fold cross-validation. 
Each value in the table is the average from three trials.
\#pLU denote the number of pLUs, and \#C denotes the final number of clusters.}
\label{tab:clustering}
\end{table*}

\paragraph{Comparison Methods}
We used BERT\footnote{\url{https://huggingface.co/bert-base-uncased}} from Hugging Face \cite{wolf-etal-2020-transformers} to obtain contextualized word embeddings.
We compared 12 methods, which comprised the vanilla model (\textbf{Vanilla}) and five fine-tuned models (\textbf{Contrastive}, \textbf{Triplet}, \textbf{Softmax}, \textbf{ArcFace}, \textbf{AdaCos}) with one-step clustering and two-step clustering.
All embeddings were processed with $l_2$ normalization.
Regarding hyperparameters, the batch size was 32, the learning rate was 1e-5, and the number of epochs for fine-tuning was five.
Also, the feature scale for ArcFace was 64.
The optimization algorithm was AdamW \cite{loshchilov2017decoupled}.

We compared our methods with the three unsupervised methods used in Subtask-A of SemEval-2019 Task 2 \cite{qasemizadeh2019} in addition to the method of \newcite{yamada-etal-2021-semantic} that corresponds to our method with the Vanilla model.\footnote{The SemEval-2019 Task 2 dataset is no longer available, as described on its official web page; thus, we excluded this dataset from the experiments.}
Regarding those three methods, first, \newcite{arefyev2019} performed group-average clustering by using the BERT embedding of a frame-evoking verb and then split each cluster into two by using TF-IDF features based on the verb's substitutes, which were generated using Hearst-like patterns \cite{hearst-1992-automatic} with BERT. 
Second, \newcite{anwar2019} performed group-average clustering through concatenation of the embedding of the frame-evoking verb and the average embedding of all words in the sentence as obtained by skip-gram.
Third, \newcite{ribeiro2019} applied graph clustering by Chinese whispers \cite{biemann2006} with the ELMo embedding of the frame-evoking verb.

\paragraph{Evaluation Metrics}
For evaluation metrics, we used the \textsc{Purity} (\textsc{Pu}), the \textsc{inverse Purity} (\textsc{iPu}), and their harmonic mean, the \textsc{F-score} (\textsc{PiF}) \cite{zhao2001}; and the \textsc{B-cubed Precision} \textsc{(BcP}), the \textsc{B-cubed Recall} (\textsc{BcR}), and their harmonic mean, the \textsc{F-score} (\textsc{BcF}) \cite{bagga1998}.
\textsc{Pu} is a metric of the degree to which a cluster is occupied by a single label, while \textsc{iPu} is a metric of the degree to which a single label is concentrated in a single cluster.
\textsc{BcP} and \textsc{BcR} evaluate the precision and recall for each sample, respectively, without associating clusters and labels.
The shared task at SemEval-2019 Task 2 ranked systems according to \textsc{BcF}.

\begin{table}[!t]
\centering
\begin{tabular}{lrr} \bhline
Method & \textsc{PiF}  & \textsc{BcF}  \\ \hline
\newcite{arefyev2019} & 65.5 & 57.4 \\ 
\newcite{anwar2019} & 62.2 & 52.2 \\ 
\newcite{ribeiro2019} & 58.2 & 46.8 \\ 
\newcite{yamada-etal-2021-semantic} & 66.9 & 56.5 \\ \hline
Ours (one-step \& Triplet) & 73.9 & 63.9 \\
Ours (one-step \& AdaCos) & 73.5 & 62.9 \\
Ours (two-step \& Triplet) & 74.8 & 66.0 \\
Ours (two-step \& AdaCos) & \textbf{75.6} & \textbf{66.2} \\ \bhline
\end{tabular}
\caption{Results of comparison with previous methods.}
\label{tab:previous}
\end{table}

\begin{table*}[t!]
\centering
\begin{tabular}{cc cc} \bhline
\multirow{2}{*}{Clustering} &\multirow{2}{*}{Model} & \textsc{PiF} & \textsc{BcF} \\
&
& \ \ \ \ \ 1 / \ \ \ \ \ 2 / \ \ \ \ \ 5 / \ \ \ 10 / \ \ all 
& \ \ \ \ \ 1 / \ \ \ \ \ 2 / \ \ \ \ \ 5 / \ \ \ 10 / \ \ all  \\ \hline

& Vanilla       & 54.9 / 54.9 / 54.9 / 54.9 / 54.9 & 42.6 / 42.6 / 42.6 / 42.6 / 42.6 \\ \cdashline{2-4}
& Contrastive   & 52.0 / 60.3 / 62.6 / 61.7 / 62.8 & 39.0 / 48.3 / 50.8 / 49.7 / 51.0 \\
One-step clustering
& Triplet       & \textbf{68.2} / \textbf{70.9} / \textbf{71.7} / \textbf{72.9} / 73.3 & \textbf{57.4} / \textbf{60.6} / \textbf{61.8} / \textbf{63.0} / \textbf{63.9} \\
{\small (group-average clustering)} 
& Softmax       & 52.9 / 54.7 / 64.9 / 69.4 / 71.0 & 39.3 / 41.7 / 52.8 / 57.8 / 59.9 \\
& ArcFace       & 58.1 / 62.5 / 69.2 / 71.9 / 73.1 & 45.6 / 50.6 / 59.1 / 62.1 / 63.3 \\
& AdaCos        & 57.5 / 59.7 / 66.5 / 70.9 / \textbf{73.5} & 44.7 / 47.1 / 54.6 / 60.0 / 62.9 \\  \hline \hline

& Vanilla       & 66.9 / 66.9 / 66.9 / 66.9 / 66.9 & 56.5 / 56.5 / 56.5 / 56.5 / 56.5 \\ \cdashline{2-4}
\multirow{2}{*}{Two-step clustering}
& Contrastive   & 68.4 / 66.3 / 67.7 / 67.4 / 65.7 & 58.4 / 56.0 / 57.7 / 57.1 / 54.8 \\
\multirow{2}{*}{{\small (X-means \&}}
& Triplet       & \textbf{71.7} / \textbf{72.5} / \textbf{73.9} / 74.0 / 74.8 & \textbf{62.4} / \textbf{63.1} / \textbf{64.8} / 64.9 / 66.0 \\
\multirow{2}{*}{\small group-average clustering)}
& Softmax       & 67.6 / 70.1 / 72.6 / 73.0 / 71.9 & 57.7 / 60.6 / 63.2 / 63.4 / 61.4 \\
& ArcFace       & 66.5 / 66.6 / 69.2 / 72.9 / 73.3 & 56.1 / 56.4 / 59.5 / 63.2 / 63.8 \\
& AdaCos        & 67.3 / 69.4 / 73.4 / \textbf{74.3} / \textbf{75.6} & 57.6 / 59.7 / 64.5 / \textbf{65.3} / \textbf{66.2} \\  \bhline

\end{tabular}
\caption{Experimental results on semantic frame induction using the vanilla and fine-tuned models over three-fold cross-validation when varying the maximum number of training instances.
Each numeric column lists the number of instances per LU included in the training set.
Each value in the table is the average from three trials.}
\label{tab:n_instances}
\end{table*}

\subsection{Results}
Table \ref{tab:clustering} summarizes the experimental results with the 12 methods.
The fine-tuned models, especially the Triplet, ArcFace, and AdaCos models, obtained higher \textsc{PiF} and \textsc{BcF} scores than the Vanilla model except for the two-step clustering method with the Contrastive model.
The reason why the Contrastive model performed worse than the other fine-tuned models could be that the space that represents the frame does not match the cluster size due to train the distance according to a fixed margin.
Table \ref{tab:previous} also lists the comparison results with previous methods.
Our fine-tuning methods achieved higher \textsc{PiF} and \textsc{BcF} scores than the previous methods.
These results indicate that fine-tuning with deep metric learning helps to improve the performance of semantic frame induction.

From Table \ref{tab:clustering}, we can see that the two-step clustering methods tended to obtain higher overall scores than the one-step clustering methods.
However, the difference in \textsc{BcF} scores between the one-step and two-step clustering methods with the Vanilla model was 13.9, whereas the difference in the maximum \textsc{BcF} scores for both clustering methods with the fine-tuned models was only 2.3.
Thus, for fine-tuning models, one-step clustering is still a good option in addition to two-step clustering.
Note that one-step clustering is more straightforward to implement than two-step clustering, but it requires more computation time\footnote{In our experiments with 16-core Intel Xeon Gold 6134 CPU at 3.20 GHz, the computation times were about 10 minutes for one-step clustering and about 5 minutes for two-step clustering.} and CPU memory to cluster many instances at once.
Regarding the weight $\alpha$, two-step clustering tended to incorporate $v_{mask}$ more than one-step clustering did in not only the Vanilla model but also the fine-tuned models.
These results suggest that two-step clustering remains effective in masking a verb's surface information even after fine-tuning.

The balance between \textsc{BcP} and \textsc{BcR} in the clustering evaluation metric depends on the final number of frame clusters, \#C in Table \ref{tab:clustering}. 
In the extreme case, \textsc{BcR} is 1 if \#C is 1, and \textsc{BcP} is 1 if \#C is equal to the number of instances. 
Hence, among models with roughly the same \textsc{BcF}, those with a fewer number of clusters tend to have higher \textsc{BcR}. 
For example, as shown in Table \ref{tab:clustering}, \#C of Triplet in two-step clustering is 454, while that of AdaCos is 656, and we can confirm that Triplet, which has fewer clusters, obtains higher \textsc{BcR} than AdaCos.

\subsection{Effect of Number of Training Instances}
We found that fine-tuned methods outperformed previous unsupervised methods when the number of training instances was around 30,000.
However, the annotation cost of building a resource like FrameNet is high, so the fewer instances used for training, the easier it is to build other language resources and apply them to other tasks.
Thus, we experimented with varying the number of training instances.
Specifically, for each LU in the training set, the maximum number of instances was varied among 1, 2, 5, 10, and all instances.
The resulting average numbers of training instances for the three sets were 1,273, 2,445, 5,680, 10,053, and 27,536, respectively.
The numbers of verbs, LUs, and frames were the same in each setting.

Table \ref{tab:n_instances} lists the \textsc{PiF} and \textsc{BcF} scores for each method.
Because the Vanilla model was not fine-tuned, its scores are the same in each setting.
The Triplet model achieved high scores even with a small number of training instances.
In the two-step clustering method with the Triplet model, the score difference between the cases of ``1'' and ``all'' is only 3.1 for \textsc{PiF} and 3.6 for \textsc{BcF}, even though the number of training instances is quite different, i.e., 1,273 vs. 27,536.
These results show that even when a small number of examples is annotated for each meaning of a verb, this method can be expected to perform considerably better than unsupervised methods.
In contrast, the Softmax, ArcFace, and AdaCos models obtained scores closer to the Triplet model in the cases of ``5'' or ``10'' but performed considerably worse with an even smaller number of training instances.
We conclude that the relatively poor performance of these models with a small number of training instances was due to insufficient training of the linear layer's weights.

\section{Analysis of Fine-tuned Embedding}
It is not easy to analyze the properties of an embedding in clustering evaluation because the performance depends on the clustering method and the number of clusters.
To better understand the fine-tuned embeddings, we performed a similarity ranking evaluation and visualized the embeddings.

\subsection{Similarity Ranking Evaluation}
We evaluated the models by ranking instances according to their embedding similarity.
Specifically, we took one verb instance as a query instance; then, we computed the cosine similarity of the embeddings between the query instance and the remaining verb instances and evaluated the similarity rankings of the instances in descending order.
We used $v_{w+m}$ with the same weight $\alpha$ that was used for the one-step clustering in Section \ref{sec:experiment}.
We chose recall as the metric to evaluate the instance distribution.
This metric computes the average matching rate between true instances, which are instances of the same frame as the query instance, and predicted instances, which are obtained by extracting the same number of top-ranked instances as the number of true instances.
For example, Set 1 of Table \ref{tab:dataset} had 153 instances of the \textsc{Filling} frame out of 28,314 total instances.
When one of these instances was the query instance, the number of true instances would be 152.
Thus, from the total instances, we would extract the top 152 instances that were similar to the query, and if 114 instances were true instances, the score would be $114/152=0.75$.

\begin{table}[t!]
\centering
\begin{tabular}{c c cc} \bhline
Model       & \textsc{All}  & \textsc{Same} & \textsc{Diff} \\ \hline
Vanilla     & 35.9          & 68.5          & 17.4 \\ \hdashline
Contrastive & 46.6          & 67.7          & 28.4 \\
Triplet     & 60.6          & 74.2          & 40.7 \\
Softmax     & 60.5          & 73.1          & 41.4 \\
ArcFace     & 58.8          & \textbf{75.6} & 37.4 \\
AdaCos      & \textbf{62.1} & 74.7          & \textbf{42.8} \\ \bhline
\end{tabular}
\caption{Experimental results of recall for the similarity ranking evaluation over three-fold cross-validation.
\textsc{All}, \textsc{Same}, and \textsc{Diff} indicate a search space of all instances, instances of the same verb, and instances of different verbs, respectively, for the query instance.}
\label{tab:ranking}
\end{table}

We performed the similarity ranking evaluation in three settings with respect to the search space of the ranked instances: \textsc{All}, which included all instances, \textsc{Same}, which included only instances of the same verb as the query, and \textsc{Diff}, which included only instances of different verbs as the query. 
Table \ref{tab:ranking} lists the results.
The results for \textsc{All} show that all of the fine-tuned models were improved over the Vanilla model; in particular, the four fine-tuned models besides the Contrastive model performed very well, improving by more than 20.
We thus confirmed that instances of the same frame were trained to be close to each other and instances of different frames are trained to be distant from each other.
Score improvements were observed for both \textsc{Same} and \textsc{Diff}, and as expected, \textsc{Same} scored higher than \textsc{Diff} both before and after fine-tuning. 
However, the improvement was much larger for \textsc{Diff} than for \textsc{Same}, suggesting that the improvement in clustering performance by fine-tuning was mainly due to the fact that different verbs evoking the same frame were trained to be close to each other.

\begin{table}[t!]
\centering
\begin{tabular}{ccc} \bhline
Model       & \textsc{Overlap}  & \textsc{Non-Overlap} \\ \hline
Vanilla     & 33.9              & 47.3 \\ \hdashline
Contrastive & 45.5              & 52.2 \\
Triplet     & 59.0              & \textbf{69.0} \\
Softmax     & 59.7              & 65.0 \\
ArcFace     & 57.1              & 68.4 \\
AdaCos      & \textbf{61.2}     & 67.0 \\ \bhline
\end{tabular}
\caption{
Separately aggregated scores for \textsc{All} in Table \ref{tab:ranking} in the \textsc{Overlap} and \textsc{Non-Overlap} cases.}
\label{tab:ranking_agg}
\end{table}

\begin{figure*}[!t]
\centering
\includegraphics[width=0.85\linewidth]{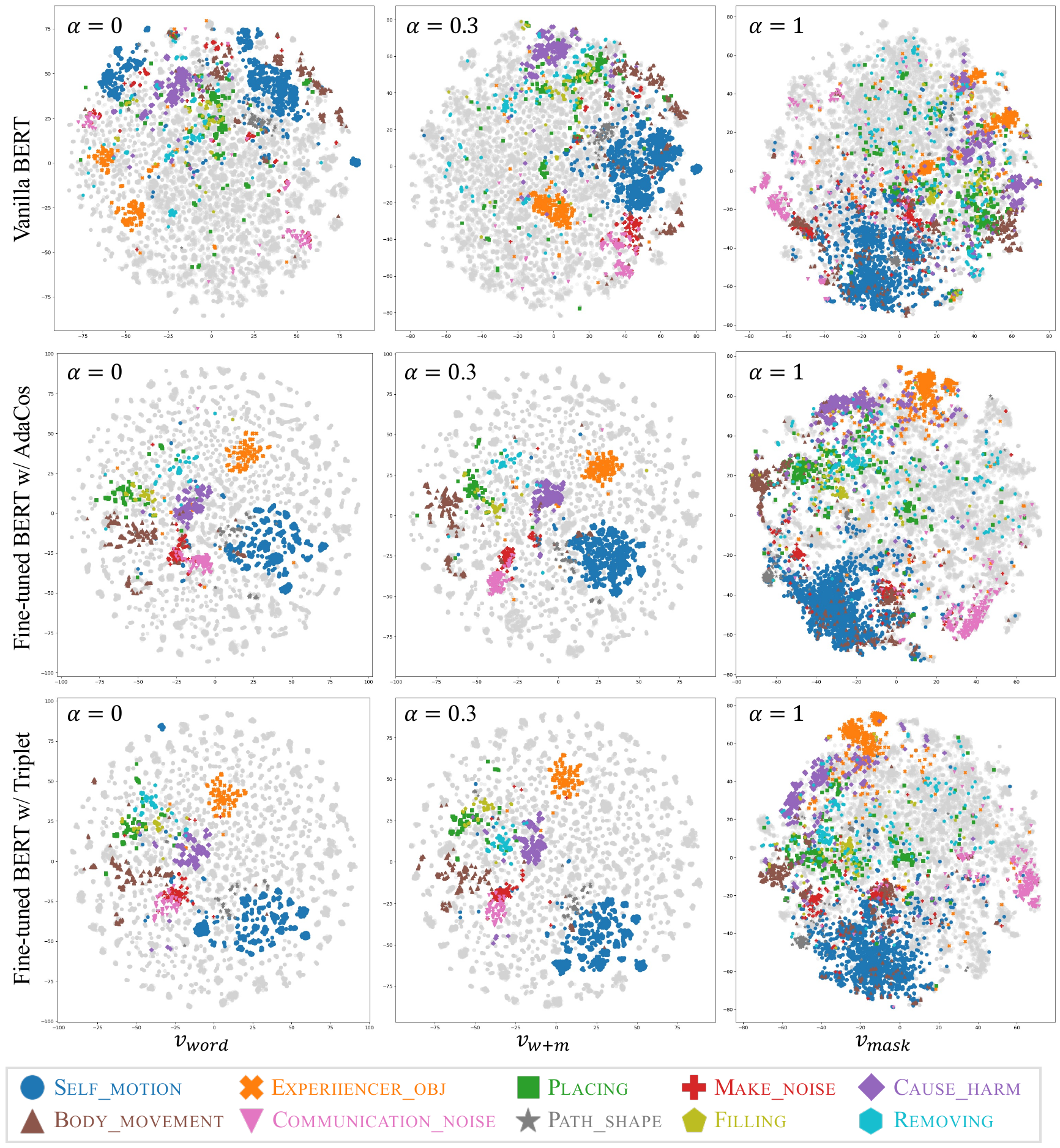}
\caption{2D t-SNE projections of $v_{word}$, $v_{w+m}$, and $v_{mask}$ for the Vanilla, AdaCos, and Triplet models, respectively, for all instances with Set 1 in Table \ref{tab:dataset} as the test set. The top 10 semantic frames with the highest numbers of instances in this set are highlighted.}
\label{fig:visualization}
\end{figure*}

It is important to further examine whether the improved performance might have resulted only from the frames included in the training set.
That is, we need to verify that the embedding of an instance of an untrained frame could be associated with a correct frame.
To investigate this, we aggregated the scores separately for cases in which the frames of the query instance were included in the training set (\textsc{Overlap}) and for cases in which they were not (\textsc{Non-Overlap}).
Table \ref{tab:ranking_agg} lists the separately aggregated results for \textsc{All} in Table \ref{tab:ranking}.
All of the fine-tuned models obtained higher scores than the Vanilla model for not only frames that were in the training set but also frames that were not.
Note that the scores for \textsc{Non-Overlap} were higher overall than those for \textsc{Overlap}.
This result may be counterintuitive, but the reason is that the frames in the \textsc{Non-Overlap} case were only evoked by a few verbs, making it relatively easy to obtain higher ranking of instances of the same frame as the query.

\subsection{Embedding Visualization}
To intuitively understand the embeddings given by the Vanilla model and two fine-tuned models, we visualized them by t-SNE.
Figure \ref{fig:visualization} shows the two-dimensional t-SNE projection of the contextualized embeddings of the frame-evoking verbs for all instances when Set 1 in Table \ref{tab:dataset} is the test set.
We used $v_{word}$, $v_{w+m}$, and $v_{mask}$ for the Vanilla, AdaCos, and Triplet models, respectively.
The weight $\alpha$ for $v_{w+m}$ was 0.3, which was the best value for one-step clustering methods with the Triplet and AdaCos models in Section \ref{sec:experiment}.\footnote{For the Vanilla model, $\alpha$ was 0, which was the same as for $v_{w+m}$ and $v_{word}$; thus, the results here are for $\alpha$ of 0.3.}
We highlight the top ten semantic frames with the highest numbers of instances in this set.

In the Vanilla model, the instances for $v_{word}$ tended to be grouped by frame but were not sufficiently grouped into clusters.
For example, the instances of the \textsc{Self\_motion} frame were divided into two large groups, while those of the \textsc{Removing} frame were scattered.
The instances for $v_{mask}$ were somewhat more scattered than those for $v_{word}$.
In addition, $v_{w+m}$ tended to group instances of the same frame.

In the AdaCos and Triplet models, the instances for $v_{word}$ were grouped much better for each frame than those for non-fine-tuned $v_{word}$.
The results also confirmed that instances of frames with similar meanings, such as the \textsc{Placing} and \textsc{Filling} frames, were both identifiable and close.
However, fine-tuned $v_{word}$ formed many lumps of instances.
This suggests that deep metric learning incorporates too much of a verb's surface information.
On the other hand, fine-tuned $v_{mask}$ was somewhat better than non-fine-tuned $v_{mask}$, but not as good as fine-tuned $v_{word}$.
As deep metric learning may require the surface information about a verb to be induced, so fine-tuned $v_{mask}$ may not work well.
The instances in fine-tuned $v_{w+m}$ were better grouped than those for fine-tuned $v_{word}$, because instances of the same frame were more grouped.

\section{Conclusion}
We worked on the supervised semantic frame induction, and we proposed a model that uses deep metric learning to fine-tune a contextualized embedding model and applied the fine-tuned contextualized embeddings to perform semantic frame induction.
In our experiments, we showed that fine-tuned BERT models with the triplet, ArcFace, and AdaCos losses are quite promising for semantic frame induction, as the human intuition in developing semantic frames such as those in FrameNet can be well captured by deep metric learning.
In particular, the fine-tuned BERT model with the triplet loss performed considerably better than vanilla BERT even when the number of training instances was small; accordingly, the fine-tuned model is expected to have a wide range of applications.
We also found that the one-step clustering can be a good choice in addition to two-step clustering when performing fine-tuning.

The ultimate goal of this study is to automatically construct semantic frame knowledge from large text corpora.
This goal requires not only grouping the verbs according to the frames that they evoke but also grouping their arguments according to the frame element roles that they fill.
Our proposed fine-tuned contextualized word embedding with deep metric learning could be effective for clustering arguments as it is for clustering verbs.
We would like to explore how to achieve this goal.

\section*{Limitations}
In this study, we only conducted experiments with English FrameNet, so it is unclear how useful this method will be for other corpora and multilingual resources.
However, since our method does not depend on the properties of the specific corpus and language, it is quite possible that fine-tuning would improve the scores in other datasets.
In addition, as our method requires supervised data from a semantic frame knowledge resource, some annotation will be necessary when applying the method to other languages that lack such a resource.

\section*{Acknowledgements}
This work was supported by JST FOREST Program, Grant Number JPMJFR216N and JSPS KAKENHI Grant Numbers 21K12012 and 22J14993.

\bibliography{eacl2023}

\begin{thebibliography}{32}
\expandafter\ifx\csname natexlab\endcsname\relax\def\natexlab#1{#1}\fi

\bibitem[{Anwar et~al.(2019)Anwar, Ustalov, Arefyev, Ponzetto, Biemann, and
  Panchenko}]{anwar2019}
Saba Anwar, Dmitry Ustalov, Nikolay Arefyev, Simone~Paolo Ponzetto, Chris
  Biemann, and Alexander Panchenko. 2019.
\newblock \href {https://www.aclweb.org/anthology/S19-2018/} {{HHMM} at
  {S}em{E}val-2019 task 2: Unsupervised frame induction using contextualized
  word embeddings}.
\newblock In \emph{Proceedings of the 13th International Workshop on Semantic
  Evaluation (SemEval 2019)}, pages 125--129.

\bibitem[{Arefyev et~al.(2019)Arefyev, Sheludko, Davletov, Kharchev,
  Nevidomsky, and Panchenko}]{arefyev2019}
Nikolay Arefyev, Boris Sheludko, Adis Davletov, Dmitry Kharchev, Alex
  Nevidomsky, and Alexander Panchenko. 2019.
\newblock \href {https://www.aclweb.org/anthology/S19-2004/} {Neural {GRANN}y
  at {S}em{E}val-2019 task 2: A combined approach for better modeling of
  semantic relationships in semantic frame induction}.
\newblock In \emph{Proceedings of the 13th International Workshop on Semantic
  Evaluation (SemEval 2019)}, pages 31--38.

\bibitem[{Bagga and Baldwin(1998)}]{bagga1998}
Amit Bagga and Breck Baldwin. 1998.
\newblock \href {https://www.aclweb.org/anthology/P98-1012/} {Entity-based
  cross-document coreferencing using the vector space model}.
\newblock In \emph{Proceedings of the 36th Annual Meeting of the Association
  for Computational Linguistics and 17th International Conference on
  Computational Linguistics (ACL-COLING 1998)}, pages 79--85.

\bibitem[{Baker et~al.(1998)Baker, Fillmore, and Lowe}]{baker1998}
Collin~F Baker, Charles~J Fillmore, and John~B Lowe. 1998.
\newblock \href {https://www.aclweb.org/anthology/P98-1013/} {The {B}erkeley
  {F}rame{N}et project}.
\newblock In \emph{Proceedings of the 36th Annual Meeting of the Association
  for Computational Linguistics and 17th International Conference on
  Computational Linguistics (ACL-COLING 1998)}, pages 86--90.

\bibitem[{Biemann(2006)}]{biemann2006}
Chris Biemann. 2006.
\newblock \href {https://www.aclweb.org/anthology/W06-3812/} {Chinese whispers:
  An efficient graph clustering algorithm and its application to natural
  language processing problems}.
\newblock In \emph{Proceedings of TextGraphs: the First Workshop on Graph Based
  Methods for Natural Language Processing (TextGraphs 2006)}, pages 73--80.

\bibitem[{Das et~al.(2014)Das, Chen, Martins, Schneider, and
  Smith}]{das-etal-2014-frame}
Dipanjan Das, Desai Chen, Andr{\'e} F.~T. Martins, Nathan Schneider, and
  Noah~A. Smith. 2014.
\newblock \href {https://aclanthology.org/J14-1002} {Frame-semantic parsing}.
\newblock \emph{Computational Linguistics}, 40(1):9--56.

\bibitem[{Deng et~al.(2019)Deng, Guo, Xue, and Zafeiriou}]{deng2019arcface}
Jiankang Deng, Jia Guo, Niannan Xue, and Stefanos Zafeiriou. 2019.
\newblock \href
  {https://openaccess.thecvf.com/content_CVPR_2019/html/Deng_ArcFace_Additive_Angular_Margin_Loss_for_Deep_Face_Recognition_CVPR_2019_paper.html}
  {Arc{F}ace: Additive angular margin loss for deep face recognition}.
\newblock In \emph{Proceedings of the IEEE/CVF Conference on Computer Vision
  and Pattern Recognition (CVPR 2019)}, pages 4690--4699.

\bibitem[{Devlin et~al.(2019)Devlin, Chang, Lee, and Toutanova}]{devlin2019}
Jacob Devlin, Ming-Wei Chang, Kenton Lee, and Kristina Toutanova. 2019.
\newblock \href {https://www.aclweb.org/anthology/N19-1423/} {{BERT}:
  Pre-training of deep bidirectional transformers for language understanding}.
\newblock In \emph{Proceedings of the 2019 Conference of the North {A}merican
  Chapter of the Association for Computational Linguistics: Human Language
  Technologies (NAACL-HLT 2019)}, pages 4171--4186.

\bibitem[{Hadsell et~al.(2006)Hadsell, Chopra, and
  LeCun}]{hadsell2006dimensionality}
Raia Hadsell, Sumit Chopra, and Yann LeCun. 2006.
\newblock \href {https://ieeexplore.ieee.org/abstract/document/1640964}
  {Dimensionality reduction by learning an invariant mapping}.
\newblock In \emph{Proceedings of the 2006 IEEE Computer Society Conference on
  Computer Vision and Pattern Recognition (CVPR 2006)}, volume~2, pages
  1735--1742.

\bibitem[{Hearst(1992)}]{hearst-1992-automatic}
Marti~A. Hearst. 1992.
\newblock \href {https://aclanthology.org/C92-2082} {Automatic acquisition of
  hyponyms from large text corpora}.
\newblock In \emph{Proceedings of the 14th International Conference on
  Computational Linguistics (COLING 1992)}.

\bibitem[{Jiang and Riloff(2021)}]{jiang-riloff-2021-exploiting}
Tianyu Jiang and Ellen Riloff. 2021.
\newblock \href {https://aclanthology.org/2021.eacl-main.206} {Exploiting
  definitions for frame identification}.
\newblock In \emph{Proceedings of the 16th Conference of the European Chapter
  of the Association for Computational Linguistics (EACL 2021)}, pages
  2429--2434.

\bibitem[{Kaya and Bilge(2019)}]{kaya2019deep}
Mahmut Kaya and Hasan~{\c{S}}akir Bilge. 2019.
\newblock \href {https://www.mdpi.com/2073-8994/11/9/1066/htm} {Deep metric
  learning: A survey}.
\newblock \emph{Symmetry}, 11(9):1066.

\bibitem[{Liu et~al.(2017)Liu, Wen, Yu, Li, Raj, and Song}]{liu2017sphereface}
Weiyang Liu, Yandong Wen, Zhiding Yu, Ming Li, Bhiksha Raj, and Le~Song. 2017.
\newblock \href
  {https://openaccess.thecvf.com/content_cvpr_2017/html/Liu_SphereFace_Deep_Hypersphere_CVPR_2017_paper.html}
  {Sphere{F}ace: Deep hypersphere embedding for face recognition}.
\newblock In \emph{Proceedings of the IEEE Conference on Computer Vision and
  Pattern Recognition (CVPR 2017)}, pages 212--220.

\bibitem[{Loshchilov and Hutter(2017)}]{loshchilov2017decoupled}
Ilya Loshchilov and Frank Hutter. 2017.
\newblock \href {https://arxiv.org/abs/1711.05101} {Decoupled weight decay
  regularization}.
\newblock In \emph{Proceedings of the 5th International Conference on Learning
  Representations (ICLR 2017)}.

\bibitem[{Maaten and Hinton(2008)}]{maaten2008}
Laurens van~der Maaten and Geoffrey Hinton. 2008.
\newblock \href {https://www.jmlr.org/papers/v9/vandermaaten08a.html}
  {Visualizing data using t-{SNE}}.
\newblock \emph{Journal of Machine Learning Research}, 9:2579--2605.

\bibitem[{Mikolov et~al.(2013)Mikolov, Sutskever, Chen, Corrado, and
  Dean}]{mikolov2013}
Tomas Mikolov, Ilya Sutskever, Kai Chen, Greg~S Corrado, and Jeff Dean. 2013.
\newblock \href
  {https://papers.nips.cc/paper/2013/hash/9aa42b31882ec039965f3c4923ce901b-Abstract.html}
  {Distributed representations of words and phrases and their
  compositionality}.
\newblock In \emph{Advances in Neural Information Processing Systems (NIPS
  2013)}, pages 3111--3119.

\bibitem[{Musgrave et~al.(2020)Musgrave, Belongie, and
  Lim}]{musgrave2020metric}
Kevin Musgrave, Serge Belongie, and Ser-Nam Lim. 2020.
\newblock \href {https://arxiv.org/pdf/2003.08505.pdf&nbsp;&hellip} {A metric
  learning reality check}.
\newblock In \emph{Proceedings of the 16th European Conference on Computer
  Vision (ECCV 2020)}, pages 681--699.

\bibitem[{Pelleg and Moore(2000)}]{pelleg2000}
Dan Pelleg and Andrew Moore. 2000.
\newblock \href
  {https://web.cs.dal.ca/~shepherd/courses/csci6403/clustering/xmeans.pdf}
  {X-means: Extending k-means with efficient estimation of the number of
  clusters}.
\newblock In \emph{Proceedings of the 17th International Conference on Machine
  Learning (ICML 2000)}, pages 727--734.

\bibitem[{Peters et~al.(2018)Peters, Neumann, Iyyer, Gardner, Clark, Lee, and
  Zettlemoyer}]{peters2018}
Matthew Peters, Mark Neumann, Mohit Iyyer, Matt Gardner, Christopher Clark,
  Kenton Lee, and Luke Zettlemoyer. 2018.
\newblock \href {https://www.aclweb.org/anthology/N18-1202/} {Deep
  contextualized word representations}.
\newblock In \emph{Proceedings of the 2018 Conference of the North {A}merican
  Chapter of the Association for Computational Linguistics: Human Language
  Technologies (NAACL-HLT 2018)}, pages 2227--2237.

\bibitem[{QasemiZadeh et~al.(2019)QasemiZadeh, Petruck, Stodden, Kallmeyer, and
  Candito}]{qasemizadeh2019}
Behrang QasemiZadeh, Miriam R.~L. Petruck, Regina Stodden, Laura Kallmeyer, and
  Marie Candito. 2019.
\newblock \href {https://www.aclweb.org/anthology/S19-2003/} {{S}em{E}val-2019
  task 2: Unsupervised lexical frame induction}.
\newblock In \emph{Proceedings of the 13th International Workshop on Semantic
  Evaluation (SemEval 2019)}, pages 16--30.

\bibitem[{Ribeiro et~al.(2019)Ribeiro, Mendon{\c{c}}a, Ribeiro, Martins~de
  Matos, Sardinha, Santos, and Coheur}]{ribeiro2019}
Eug{\'e}nio Ribeiro, V{\^a}nia Mendon{\c{c}}a, Ricardo Ribeiro, David
  Martins~de Matos, Alberto Sardinha, Ana~L{\'u}cia Santos, and Lu{\'\i}sa
  Coheur. 2019.
\newblock \href {https://www.aclweb.org/anthology/S19-2019/}
  {{L}2{F}/{INESC}-{ID} at {S}em{E}val-2019 task 2: Unsupervised lexical
  semantic frame induction using contextualized word representations}.
\newblock In \emph{Proceedings of the 13th International Workshop on Semantic
  Evaluation (SemEval 2019)}, pages 130--136.

\bibitem[{Ruppenhofer et~al.(2016)Ruppenhofer, Ellsworth, Schwarzer-Petruck,
  Johnson, and Scheffczyk}]{ruppenhofer2016}
Josef Ruppenhofer, Michael Ellsworth, Myriam Schwarzer-Petruck, Christopher~R
  Johnson, and Jan Scheffczyk. 2016.
\newblock \href {https://framenet2.icsi.berkeley.edu/docs/r1.7/book.pdf}
  {\emph{FrameNet II: Extended theory and practice}}.
\newblock International Computer Science Institute.

\bibitem[{Su et~al.(2021)Su, Li, Li, Pan, Zhang, Chai, and
  Han}]{su-etal-2021-knowledge}
Xuefeng Su, Ru~Li, Xiaoli Li, Jeff~Z. Pan, Hu~Zhang, Qinghua Chai, and Xiaoqi
  Han. 2021.
\newblock \href {https://aclanthology.org/2021.acl-long.407} {A
  knowledge-guided framework for frame identification}.
\newblock In \emph{Proceedings of the 59th Annual Meeting of the Association
  for Computational Linguistics and the 11th International Joint Conference on
  Natural Language Processing (ACL-IJCNLP 2021)}, pages 5230--5240.

\bibitem[{Swayamdipta et~al.(2017)Swayamdipta, Thomson, Dyer, and
  Smith}]{swayamdipta2017frame}
Swabha Swayamdipta, Sam Thomson, Chris Dyer, and Noah~A Smith. 2017.
\newblock \href {https://arxiv.org/abs/1706.09528} {Frame-semantic parsing with
  softmax-margin segmental {RNN}s and a syntactic scaffold}.
\newblock ar{X}iv preprint ar{X}iv:1706.09528.

\bibitem[{Wang et~al.(2018)Wang, Wang, Zhou, Ji, Gong, Zhou, Li, and
  Liu}]{wang2018cosface}
Hao Wang, Yitong Wang, Zheng Zhou, Xing Ji, Dihong Gong, Jingchao Zhou, Zhifeng
  Li, and Wei Liu. 2018.
\newblock \href
  {https://openaccess.thecvf.com/content_cvpr_2018/html/Wang_CosFace_Large_Margin_CVPR_2018_paper.html}
  {Cos{F}ace: Large margin cosine loss for deep face recognition}.
\newblock In \emph{Proceedings of the IEEE Conference on Computer Vision and
  Pattern Recognition (CVPR 2018)}, pages 5265--5274.

\bibitem[{Weinberger and Saul(2009)}]{weinberger2009distance}
Kilian~Q Weinberger and Lawrence~K Saul. 2009.
\newblock \href
  {https://www.jmlr.org/papers/volume10/weinberger09a/weinberger09a.pdf}
  {Distance metric learning for large margin nearest neighbor classification}.
\newblock \emph{Journal of Machine Learning Research}, 10(2).

\bibitem[{Wolf et~al.(2020)Wolf, Debut, Sanh, Chaumond, Delangue, Moi, Cistac,
  Rault, Louf, Funtowicz, Davison, Shleifer, von Platen, Ma, Jernite, Plu, Xu,
  Scao, Gugger, Drame, Lhoest, and Rush}]{wolf-etal-2020-transformers}
Thomas Wolf, Lysandre Debut, Victor Sanh, Julien Chaumond, Clement Delangue,
  Anthony Moi, Pierric Cistac, Tim Rault, Rémi Louf, Morgan Funtowicz, Joe
  Davison, Sam Shleifer, Patrick von Platen, Clara Ma, Yacine Jernite, Julien
  Plu, Canwen Xu, Teven~Le Scao, Sylvain Gugger, Mariama Drame, Quentin Lhoest,
  and Alexander~M. Rush. 2020.
\newblock \href {https://www.aclweb.org/anthology/2020.emnlp-demos.6}
  {Transformers: State-of-the-art natural language processing}.
\newblock In \emph{Proceedings of the 2020 Conference on Empirical Methods in
  Natural Language Processing: System Demonstrations (EMNLP 2020)}, pages
  38--45.

\bibitem[{Yamada et~al.(2021{\natexlab{a}})Yamada, Sasano, and
  Takeda}]{yamada-etal-2021-semantic}
Kosuke Yamada, Ryohei Sasano, and Koichi Takeda. 2021{\natexlab{a}}.
\newblock \href {https://aclanthology.org/2021.acl-short.102/} {Semantic frame
  induction using masked word embeddings and two-step clustering}.
\newblock In \emph{Proceedings of the 59th Annual Meeting of the Association
  for Computational Linguistics and the 11th International Joint Conference on
  Natural Language Processing (ACL-IJCNLP 2021)}, pages 811--816.

\bibitem[{Yamada et~al.(2021{\natexlab{b}})Yamada, Sasano, and
  Takeda}]{yamada-etal-2021-verb}
Kosuke Yamada, Ryohei Sasano, and Koichi Takeda. 2021{\natexlab{b}}.
\newblock \href {https://aclanthology.org/2021.findings-acl.381} {Verb sense
  clustering using contextualized word representations for semantic frame
  induction}.
\newblock In \emph{Findings of the Association for Computational Linguistics:
  ACL-IJCNLP 2021 (ACL-IJCNLP 2021 Findings)}, pages 4353--4362.

\bibitem[{Yong and Torrent(2020)}]{yong2020semi}
Zheng~Xin Yong and Tiago~Timponi Torrent. 2020.
\newblock \href {https://aclanthology.org/2020.lrec-1.431/} {Semi-supervised
  deep embedded clustering with anomaly detection for semantic frame
  induction}.
\newblock In \emph{Proceedings of the 12th Language Resources and Evaluation
  Conference (LREC 2020)}, pages 3509--3519.

\bibitem[{Zhang et~al.(2019)Zhang, Zhao, Qiao, Wang, and Li}]{zhang2019adacos}
Xiao Zhang, Rui Zhao, Yu~Qiao, Xiaogang Wang, and Hongsheng Li. 2019.
\newblock \href
  {https://openaccess.thecvf.com/content_CVPR_2019/html/Zhang_AdaCos_Adaptively_Scaling_Cosine_Logits_for_Effectively_Learning_Deep_Face_CVPR_2019_paper.html}
  {Ada{C}os: Adaptively scaling cosine logits for effectively learning deep
  face representations}.
\newblock In \emph{Proceedings of the IEEE/CVF Conference on Computer Vision
  and Pattern Recognition (CVPR 2019)}, pages 10823--10832.

\bibitem[{Zhao and Karypis(2001)}]{zhao2001}
Ying Zhao and George Karypis. 2001.
\newblock \href {https://hdl.handle.net/11299/21549} {Criterion functions for
  document clustering: Experiments and analysis}.
\newblock Technical report, Retrieved from the University of Minnesota Digital
  Conservancy.

\end{thebibliography}
\bibliographystyle{acl_natbib}

\appendix
\begin{table*}[t!]
\centering
\begin{tabular}{cccrccc}
\bhline
Clustering & Model & $\alpha$ 
& \multicolumn{1}{c}{\#pLU} & \#C & \textsc{Pu} / \textsc{iPu} / \textsc{PiF} 
& \textsc{BcP} / \textsc{BcR} / \textsc{BcF} \\ \hline
 
& Vanilla         & 0.00 & --     & 429 & 53.0 / 57.0 / 54.9 & 40.8 / 44.6 / 42.6 \\ \cdashline{2-7}
& Contrastive     & 0.00 & --     & 421 & 55.4 / 69.6 / 61.7 & 43.7 / 58.2 / 49.9 \\
One-step clustering
& Triplet        & 0.00 & --     & 424 & \textbf{69.9} / 76.6 / \textbf{73.1} & 60.3 / 67.7 / \textbf{63.7} \\
{\small (group-average clustering)} 
& Softmax        & 0.00 & --     & 433 & 65.1 / 77.4 / 70.7 & 53.3 / 67.9 / 59.7 \\
& ArcFace         & 0.00 & --     & 430 & 69.8 / 75.1 / 72.3 & \textbf{62.7} / 66.1 / 62.7 \\
& AdaCos          & 0.00 & --     & 434 & 68.2 / \textbf{77.6} / 72.6 & 62.0 / \textbf{68.1} / 62.0 \\ \hline \hline

& Vanilla       & 0.00 & -- & 429 & 53.0 / 57.0 / 54.9 & 40.8 / 44.6 / 42.6 \\ \cdashline{2-7}
& Contrastive   & 0.13 & -- & 443 & 56.9 / 70.0 / 62.8 & 45.1 / 58.6 / 51.0 \\
One-step clustering
& Triplet       & 0.23 & -- & 425 & 70.0 / 77.0 / 73.3 & \textbf{60.3} / 68.1 / \textbf{63.9} \\
{\small (group-average clustering)} 
& Softmax       & 0.23 & -- & 440 & 65.1 / 58.0 / 71.0 & 53.3 / 68.6 / 59.9 \\
& ArcFace       & 0.37 & -- & 436 & \textbf{70.3} / 76.2 / 73.1 & 59.7 / 67.4 / 63.3 \\
& AdaCos        & 0.30 & -- & 446 & 69.0 / \textbf{78.7} / \textbf{73.5} & 57.5 / \textbf{69.5} / 62.9 \\ \hline \hline

& Vanilla         & 1.00 & --  & 430 & 29.9 / 38.9 / 33.8 & 18.0 / 24.4 / 20.7 \\ \cdashline{2-7}
& Contrastive  & 1.00 & --     & 449 & 26.1 / 40.8 / 31.8 & 14.3 / 26.2 / 18.4 \\
One-step clustering
& Triplet      & 1.00 & --    & 442 & \textbf{35.5} / 47.9 / \textbf{40.8} & \textbf{21.0} / 31.8 / \textbf{25.2} \\
{\small (group-average clustering)} 
& Softmax      & 1.00 & --    & 447 & 26.1 / \textbf{50.6} / 37.4 & 15.5 / \textbf{35.3} / 21.5 \\
& ArcFace      & 1.00 & --     & 452 & 35.0 / 46.2 / 40.1 & 20.9 / 30.6 / 24.8 \\
& AdaCos       & 1.00 & --     & 432 & 33.7 / 48.0 / 39.1 & 18.3 / 32.1 / 23.3 \\ \bhline
\end{tabular}
\caption{Results on semantic frame induction from using $v_{word}$, $v_{w+m}$, and $v_{mask}$ with in one-step clustering.}
\label{fig:one-step}
\end{table*}

\begin{table*}[t!]
\centering
\begin{tabular}{cccrccc}
\bhline
Clustering & Model & $\alpha$ 
& \multicolumn{1}{c}{\#pLU} & \#C & \textsc{Pu} / \textsc{iPu} / \textsc{PiF} 
& \textsc{BcP} / \textsc{BcR} / \textsc{BcF} \\ \hline
& Vanilla         & 0.00 & 880    & 438 & 52.5 / 68.2 / 59.3 & 43.0 / 58.1 / 49.4 \\ \cdashline{2-7}
\multirow{2}{*}{Two-step clustering}
& Contrastive     & 0.00 & 2269   & 720 & 68.2 / 62.8 / 65.3 & 58.7 / 50.6 / 54.2 \\
\multirow{2}{*}{{\small (X-means \&}}
& Triplet        & 0.00 & 1277   & 500 & 73.3 / 71.0 / 71.8 & 64.8 / 60.8 / 62.2 \\
\multirow{2}{*}{\small group-average clustering)}
& Softmax        & 0.00 & 1909   & 1107 & \textbf{85.1} / 60.0 / 70.3 & \textbf{78.7} / 47.8 / 59.3 \\
& ArcFace        & 0.00 & 1139   & 438 & 70.9 / \textbf{73.3} / 72.0 & 61.2 / \textbf{63.8} / 62.4 \\
& AdaCos          & 0.00 & 1664   & 897 & 83.6 / 64.7 / \textbf{72.7} & 76.9 / 53.3 / \textbf{62.6} \\ \hline \hline

& Vanilla       & 0.67 & 877    & 444 & 60.6 / 74.9 / 66.9 & 49.7 / 65.8 / 56.5 \\ \cdashline{2-7}
\multirow{2}{*}{Two-step clustering}
& Contrastive   & 0.23 & 1904   & 689 & 69.2 / 62.5 / 65.7 & 59.5 / 50.9 / 54.8 \\
\multirow{2}{*}{{\small (X-means \&}}
& Triplet       & 0.50 & 1014   & 454 & 73.4 / \textbf{76.7} / 74.8 & 64.6 / \textbf{68.0} / 66.0 \\
\multirow{2}{*}{\small group-average clustering)}
& Softmax       & 0.43 & 1428   & 919 & \textbf{84.7} / 62.5 / 71.9 & \textbf{78.4} / 50.4 / 61.4 \\
& ArcFace       & 0.47 & 955    & 452 & 70.5 / 76.5 / 73.3 & 60.8 / 67.7 / 63.8 \\
& AdaCos        & 0.50 & 1128   & 656 & 80.8 / 71.3 / \textbf{75.6} & 73.2 / 60.9 / \textbf{66.2} \\ \hline \hline

& Vanilla         & 1.00 & 873    & 469 & 59.0 / 72.9 / 65.2 & 48.8 / 63.6 / 55.1 \\ \cdashline{2-7}
\multirow{2}{*}{Two-step clustering}
& Contrastive     & 1.00 & 2050   & 718 & 49.3 / 49.2 / 49.2 & 37.7 / 36.3 / 36.9 \\
\multirow{2}{*}{{\small (X-means \&}}
& Triplet        & 1.00 & 930    & 463 & 65.2 / \textbf{73.9} / \textbf{69.0} & 55.2 / \textbf{64.3} / \textbf{59.1} \\
\multirow{2}{*}{\small group-average clustering)}
& Softmax        & 1.00 & 1573   & 873 & \textbf{68.6} / 55.3 / 61.2 & \textbf{58.6} / 43.5 / 49.9 \\
& ArcFace        & 1.00 & 906    & 467 & 62.5 / 73.7 / 67.5 & 52.3 / \textbf{64.3} / 57.4 \\
& AdaCos          & 1.00 & 1113   & 522 & 65.4 / 65.5 / 65.2 & 55.2 / 55.2 / 54.5 \\ \bhline
\end{tabular}
\caption{Results on semantic frame induction from using $v_{word}$, $v_{w+m}$, and $v_{mask}$ with two-step clustering.}
\label{fig:two-step}
\end{table*}

\begin{table*}[t!]
\centering
\begin{tabular}{cc crccc} \bhline
Clustering & Model & $\alpha$ 
 & \multicolumn{1}{c}{\#pLU} & \#C & \textsc{Pu} / \textsc{iPu} / \textsc{PiF} 
 & \textsc{BcP} / \textsc{BcR} / \textsc{BcF} \\ \hline

& Vanilla       & 0.00 & -- & 433 & 53.1 / 57.1 / 55.0 & 40.9 / 44.7 / 42.7 \\ \cdashline{2-7}
& Contrastive   & 0.13 & -- & 433 & 57.7 / 70.7 / 63.5 & 45.2 / 59.4 / 51.3 \\
One-step clustering
& Triplet      & 0.37 & -- & 433 & \textbf{71.7} / 76.9 / \textbf{74.2} & \textbf{62.0} / 67.4 / \textbf{64.6} \\
{\small (group-average clustering)} 
& Softmax      & 0.27 & -- & 433 & 66.4 / 77.6 / 71.5 & 53.7 / 68.2 / 60.1 \\
& ArcFace       & 0.20 & -- & 433 & 70.8 / 76.1 / 73.3 & 60.6 / 67.3 / 63.8 \\
& AdaCos        & 0.37 & -- & 433 & 69.1 / \textbf{78.5} / 73.5 & 57.2 / \textbf{69.3} / 62.7 \\ \hline \hline

& Vanilla       & 0.67 & 877    & 441 & 61.3 / 75.9 / 67.7 & 50.3 / 67.0 / 57.4 \\ \cdashline{2-7}
\multirow{2}{*}{Two-step clustering}
& Contrastive  & 0.23 & 1879   & 736 & 70.3 / 63.5 / 66.6 & 60.9 / 51.8 / 55.9 \\
\multirow{2}{*}{{\small (X-means \&}}
& Triplet      & 0.50 & 1016   & 431 & 74.7 / \textbf{78.1} / 76.4 & 65.7 / \textbf{69.6} / \textbf{67.6} \\
\multirow{2}{*}{\small group-average clustering)}
& Softmax      & 0.43 & 1442   & 922 & \textbf{84.6} / 63.2 / 72.3 & \textbf{78.1} / 51.0 / 61.7 \\
& ArcFace       & 0.47 & 966    & 453 & 71.5 / 76.9 / 74.0 & 62.1 / 68.0 / 64.9 \\
& AdaCos        & 0.50 & 1120   & 606 & 79.7 / 73.8 / \textbf{76.5} & 71.7 / 63.7 / 67.3 \\ \bhline
\end{tabular}
\caption{Experimental results for semantic frame induction on the development set.}
\label{tab:clustering_dev}
\end{table*}

\begin{figure*}[!t]
\centering
\includegraphics[width=0.85\linewidth]{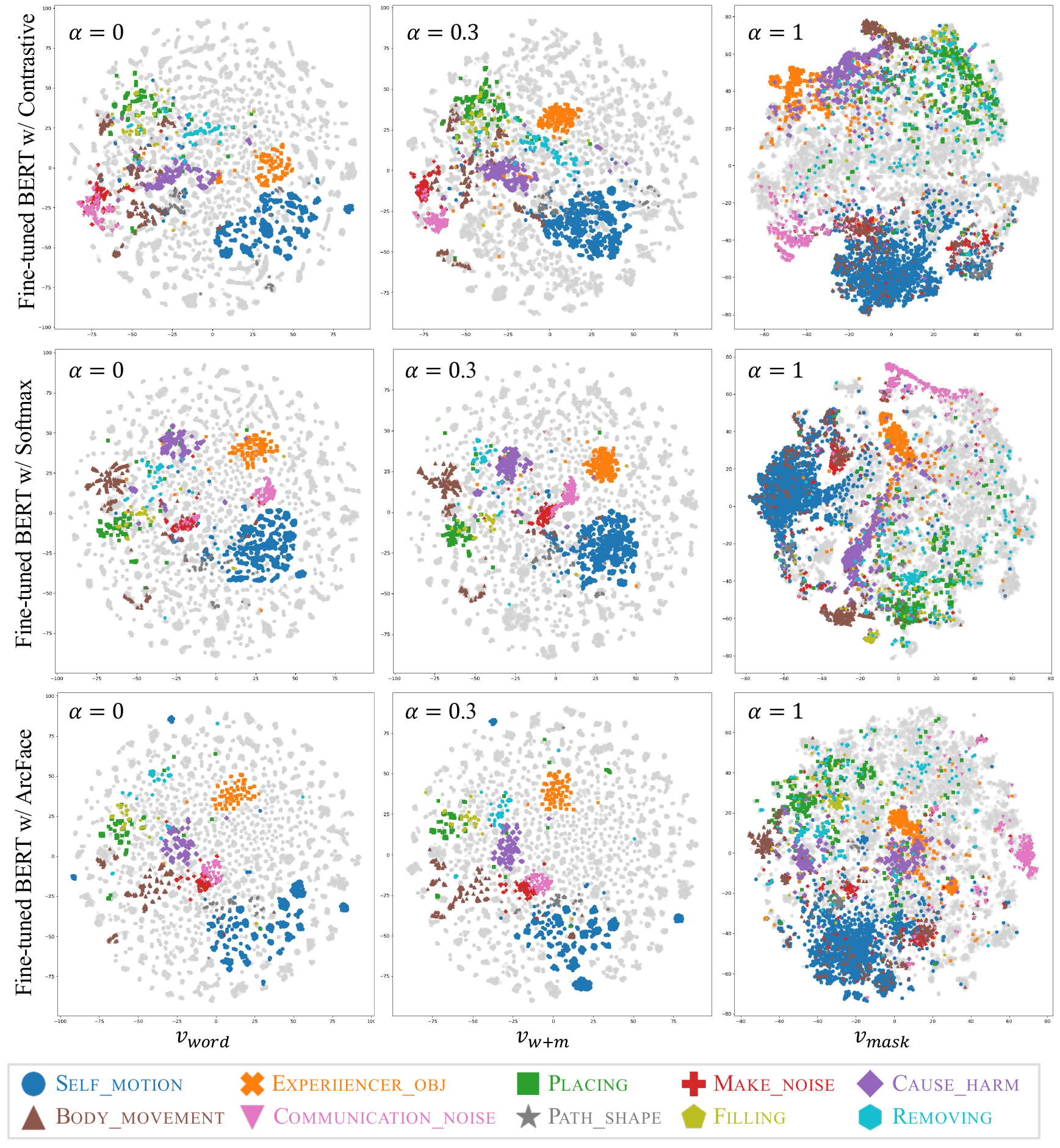}
\caption{2D t-SNE projections of $v_{word}$, $v_{w+m}$, and $v_{mask}$ for the Contrastive, Softmax, and ArcFace models, respectively, for all instances with Set 1 in Table \ref{tab:dataset} as the test set. The top 10 semantic frames with the highest numbers of instances in this set are highlighted.}
\label{fig:visualization2}
\end{figure*}

\section{Clustering Results with and without Linear Completion}
Tables \ref{fig:one-step} and \ref{fig:two-step} list our experimental results for semantic frame induction when using $v_{word}$, $v_{w+m}$, and $v_{mask}$ in one-step and two-step clustering, respectively.
The results show that $v_{w+m}$ tended to perform better than $v_{word}$ and $v_{mask}$, thus demonstraiting the usefulness of linear completion.
This tendency was noticeable for two-step clustering but more limited for one-step clustering.

Regarding the results for $v_{word}$ and $v_{mask}$, the fine-tuning was effective for $v_{word}$, as the scores improved considerably, but the effectiveness was limited for $v_{mask}$.
This was probably because the embedding of the special token ``[MASK],'' which was the source of the contextualized word embedding, was shared by all instances.

\section{Results for Semantic Frame Induction on Development Set}
Table \ref{tab:clustering_dev} lists our experimental results for semantic frame induction on the development set.
As with the test set, the fine-tuned models obtained higher \textsc{PiF} and \textsc{BcF} scores than the Vanilla model except for two-step clustering with the Contrastive model.
In particular, the Triplet, ArcFace, and AdaCos models obtained high scores for both one-step and two-step clusterings.

\section{Embedding Visualization of Remaining Models}
In Figure \ref{fig:visualization}, we showed a two-dimensional t-SNE projection of $v_{word}$, $v_{w+m}$, and $v_{mask}$ for the Vanilla, AdaCos, and Triplet models, respectively.
Figure \ref{fig:visualization2} shows a two-dimensional t-SNE projection of  $v_{word}$, $v_{w+m}$, and $v_{mask}$ for the remaining models not included in Figure \ref{fig:visualization}, namely, the Contrastive, Softmax, and ArcFace models, with the same setting.
Figure \ref{fig:visualization2} confirms that the three fine-tuned models, as well as the two fine-tuned models shown in Figure \ref{fig:visualization}, are more coherent semantically than the Vanilla model, and the tendency of $v_{word}$, $v_{w+m}$, and $v_{mask}$ is similar.
In addition, for the Contrastive model, whose performance was relatively poor among the fine-tuning models in Table 3, it was confirmed that the instances were somewhat scattered compared to the Softmax and ArcFace models.
For example, the instances of the \textsc{Body\_movement} frame were scattered.
This result confirmed the consistency of scores and visualization.
The results show consistency between scores and visualization.

\end{document}